\documentclass[letterpaper]{article} 
\usepackage{aaai19}  
\usepackage{times}  
\usepackage{courier}  
\usepackage{url}  
\usepackage{graphicx}  
\usepackage{amsmath}
\usepackage{booktabs}
\usepackage{multirow}
\usepackage{algorithm}
\usepackage{algorithmic}
\usepackage{amssymb}
\usepackage{graphicx}
\usepackage{bbm}
\usepackage{color}
\usepackage{xspace}
\usepackage[scaled=.90]{helvet} 
\usepackage{subfigure}
\usepackage{capt-of}
\usepackage{hyperref}

\newcommand{\ie}{\emph{i.e.,}\xspace}
\newcommand{\eg}{\emph{e.g.,}\xspace}

\newcommand{\citenoun}[1]{\citeauthor{#1}~\shortcite{#1}}

\begin{document}
%
\title{Logic Attention Based Neighborhood Aggregation for Inductive \\Knowledge Graph Embedding}
\author{
Peifeng Wang$^{1}$\footnotemark[1], Jialong Han$^{2}$, Chenliang Li$^{3}$, Rong Pan$^{1}$\thanks{This work was done during Peifeng Wang's internship at Tencent AI Lab. The corresponding author is Rong Pan.}\\ 
$^{1}$Sun Yat-sen University, China, $^{2}$Tencent AI Lab, China, $^{3}$Wuhan University, China \\
$^{1}$\{wangpf3@mail2., panr@\}sysu.edu.cn, $^{2}$jialonghan@gmail.com, $^{3}$cllee@whu.edu.cn 
}
\maketitle
\begin{abstract}
Knowledge graph embedding aims at modeling entities and relations with low-dimensional vectors. Most previous methods require that all entities should be seen during training, which is unpractical for real-world knowledge graphs with new entities emerging on a daily basis. Recent efforts on this issue suggest training a neighborhood aggregator in conjunction with the conventional entity and relation embeddings, which may help embed new entities inductively via their existing neighbors. However, their neighborhood aggregators neglect the unordered and unequal natures of an entity's neighbors. To this end, we summarize the desired properties that may lead to effective neighborhood aggregators. We also introduce a novel aggregator, namely, Logic Attention Network (LAN), which addresses the properties by aggregating neighbors with both rules- and network-based attention weights. By comparing with conventional aggregators on two knowledge graph completion tasks, we experimentally validate LAN's superiority in terms of the desired properties. The code is available at \url{https://github.com/wangpf3/LAN}.

\end{abstract}

\section{Introduction}
Knowledge graphs (KGs) such as Freebase~\cite{bollacker2008freebase}, DBpedia~\cite{auer2007dbpedia}, and YAGO~\cite{mahdisoltani2014yago3} play a critical role in various NLP tasks, including question answering~\cite{hao2017end}, information retrieval~\cite{xiong2015esdrank}, and personalized recommendation~\cite{zhang2016collaborative}.
A typical KG consists of numerous facts about a predefined set of entities.
Each fact is in the form of a triplet $(subject,relation,object)$ (or $(s,r,o)$ for short), where $s$ and $o$ are two entities and $r$ is a relation the fact describes.
Due to the discrete and incomplete natures of KGs, various KG embedding models are proposed to facilitate KG completion tasks, \eg link prediction and triplet classification.
After vectorizing entities and relations in a low-dimensional space, those models predict missing facts by manipulating the involved entity and relation embeddings.

\begin{figure}[h]
	\centering
 	\includegraphics[scale=0.485]{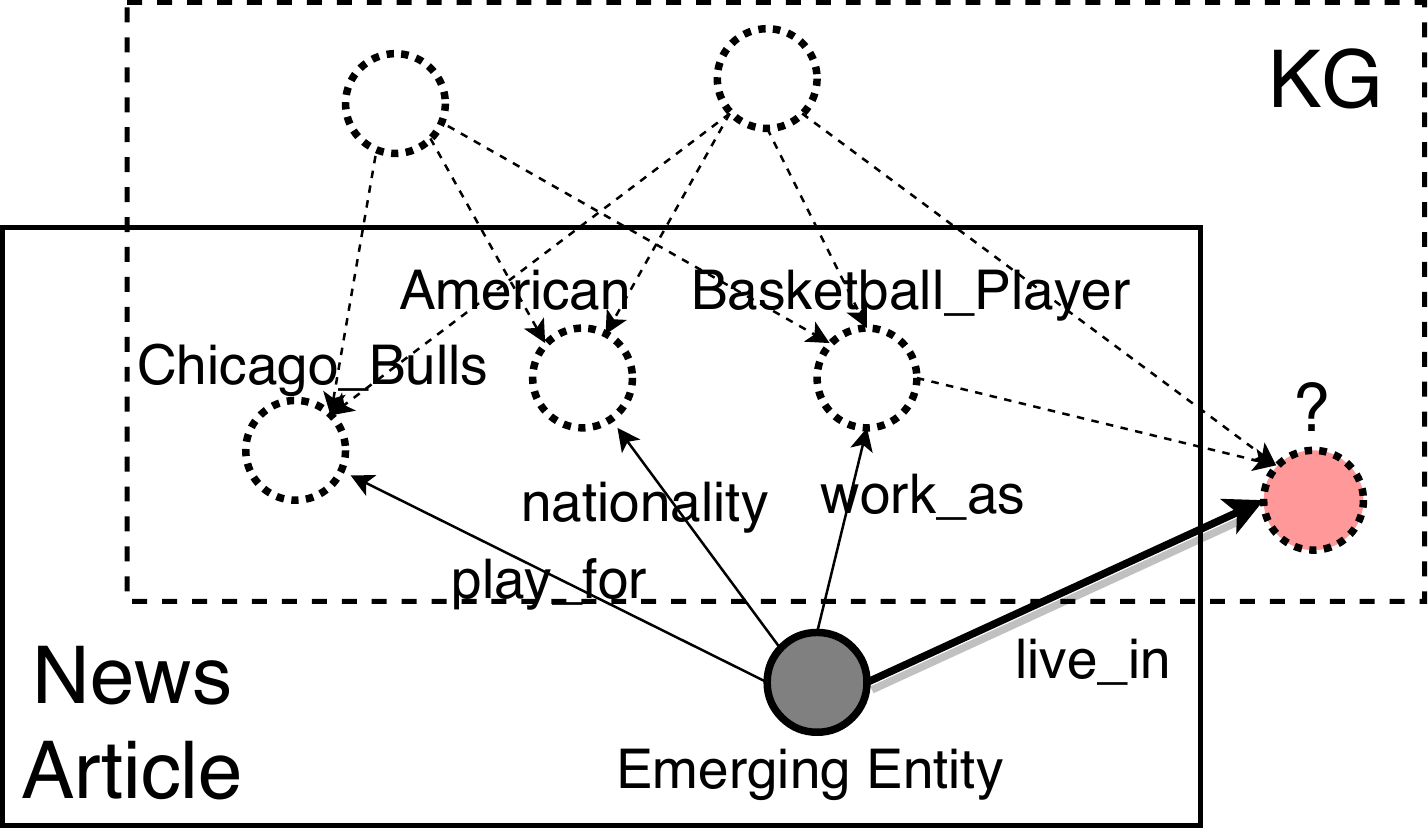}
	\caption{A motivating example of emerging KG entities. Dotted circles and arrows represent the existing KG while solid ones are brought by the emerging entity.}
	\label{fig:mj}
\end{figure}


Although proving successful in previous studies, traditional KG embedding models simply ignore the \emph{evolving} nature of KGs.
They require all entities to be present when training the embeddings.
However, \citenoun{shi2018open} suggest that, on DBpedia, 200 new entities emerge on a daily basis between late 2015 and early 2016.
Given the infeasibility of retraining embeddings from scratch whenever new entities come, missing facts about emerging entities are, unfortunately, not guaranteed to be inferred in time.

By transforming realistic networks, \eg citation graphs, social networks, and protein interaction graphs, to simple graphs with single-typed and undirected edges, recent explorations~\cite{hamilton2017representation} shed light on the evolution issue for homogeneous graphs.
While learning embeddings for existing nodes, they inductively learn a \emph{neighborhood aggregator} that represents a node by aggregating its neighbors' embeddings.
The embeddings of unseen nodes can then be obtained by applying the aggregator on their existing neighbors.

It is well received that KGs differ from homogeneous graphs by their multi-relational structure~\cite{shi2017survey}.
Despite the difference, it seems promising to generalize the neighborhood aggregating scheme to embed emerging KG entities in an inductive manner.
For example, in Figure~\ref{fig:mj}, a news article may describe an emerging entity (marked gray) as well as some facts involving existing entities.
By generalizing structural information in the underlying KG, \eg other entities residing in a similar neighborhood or involving similar relations, to the current entity's neighborhood, we can infer that it may probably live in \textsf{Chicago}.

Inspired by the above example, the inductive KG embedding problem boils down to designing a KG-specific neighborhood aggregator to capture essential neighborhood information.
Intuitively, an ideal aggregator should have the following desired properties:


\begin{itemize}
	\item \textbf{Permutation Invariant} - Unlike words in text or pixels in an image, neighbors of an entity are naturally unordered. When having neighbors like \textsf{Chicago\underline{ }Bulls} and \textsf{American} as input, the aggregator should be irrelevant to potential permutations of the neighbors.  
	\item \textbf{Redundancy Aware} - Facts in KGs tend to depend on each other. \emph{E.g.}, the fact that one plays for \textsf{Chicago\underline{ }Bulls} always implies that she/he is a \textsf{Basketball\underline{ }Player}. It is beneficial to exploit such redundancy in an entity's neighborhood to make the aggregation informative.    
	\item \textbf{Query Relation Aware} - For common KG completion tasks, the query relation in concern, \eg ``live in'', is given beforehand. An aggregator may exploit such information to concentrate on relevant facts in the neighborhood, \eg ``play for'' \textsf{Chicago\underline{ }Bulls}.
\end{itemize}

This paper concentrates on KG-specific neighborhood aggregators, which is of practical importance but only received limited focus~\cite{ijcai2017-250}.
To the best of our knowledge, neither conventional aggregators for homogeneous graphs nor those for KGs satisfy all the above three properties.
In this regard, we employ the attention mechanism~\cite{bahdanau2014neural} and propose an aggregator called Logic Attention Network (LAN).
Aggregating neighbors by a weighted combination of their transformed embeddings, LAN is inherently permutation invariant.
To estimate the attention weights in LAN, we adopt two mechanisms to model relation- and neighbor-level information in a coarse-to-fine manner,
At both levels, LAN is made aware of both neighborhood redundancy and query relation.


To summarize, our contributions are:
(1) We propose three desired properties that decent neighborhood aggregators for KGs should possess.
(2) We propose a novel aggregator, \ie Logic Attention Network, to facilitate inductive KG embedding.
(3) We conduct extensive comparisons with conventional aggregators on two KG completions tasks. The results validate the superiority of LAN w.r.t.\ the three properties.

\section{Related Works}

\subsection{Transductive Embedding Models}
In recent years, representation learning problems on KGs have received much attention due to the wide applications of the resultant entity and relation embeddings. 
Typical KG embedding models include TransE~\cite{bordes2013translating}, Distmult~\cite{yang2014embedding}, Complex~\cite{trouillon2016complex}, Analogy~\cite{liu2017analogical}, to name a few. 
For more explorations, we refer readers to an extensive survey~\cite{wang2017knowledge}.
However, conventional approaches on KG embedding work in a transductive manner.
They require that all entities should be seen during training.
Such limitation hinders them from efficiently generalizing to emerging entities.

\subsection{Inductive Embedding Models}

To relieve the issue of emerging entities, several inductive KG embedding models are proposed, including ~\citenoun{xie2016representation}, ~\citenoun{shi2018open} and ~\citenoun{xie2016image} which use description text or images as inputs.
Although the resultant embeddings may be utilized for KG completion, it is not clear whether the embeddings are powerful enough to infer implicit or new facts beyond those expressed in the text/image.
Moreover, when domain experts are recruited to introduce new entities via partial facts rather than text or images, those approaches may not help much.

In light of the above scenario, existing neighbors of an emerging entity are considered as another type of input for inductive models. In~\citenoun{ijcai2017-250}, the authors propose applying Graph Neural Network (GNN) on the KG, which generates the embedding of a new entity by aggregating all its known neighbors. However, their model aggregates the neighbors via simple pooling functions, which neglects the difference among the neighbors. 
Other works like \citenoun{fu2017hin2vec} and \citenoun{tang2015pte} aim at embedding nodes for node classification given the entire graph and thus are inapplicable for inductive KG-specific tasks.
\citenoun{schlichtkrull2017modeling} and \citenoun{xiong2018one} also rely on neighborhood structures to embed entities, but they either work transductively or focus on emerging relations.

Finally, we note another related line of studies on node representation learning for homogeneous graphs.
Similar to text- or image-based inductive models for KGs, \citenoun{duran2017learning}, \citenoun{yang2016revisiting}, \citenoun{velivckovic2017graph} and~\citenoun{rossi2018deep} exploit additional node attributes to embed unseen nodes.
Another work more related to ours is~\citenoun{hamilton2017inductive}.
They tackle inductive node embedding by the neighborhood aggregation scheme.
Their aggregators either trivially treat neighbors equally or unnecessarily require them to be ordered.
Moreover, like all embedding models for homogeneous graphs, their model cannot be directly applied to KGs with multi-relational edges.

\section{Preliminaries}
\subsection{Notations}
Let $\mathcal{E}$ and $\mathcal{R}$ be two sets of entities and relations of size $n$ and $m$, respectively.
A knowledge graph is composed of a set of triplet facts, namely
\begin{equation}
\mathcal{K}=\{(s,r,o)|s\in\mathcal{E},r\in\mathcal{R},o\in\mathcal{E}\}\text{.}
\end{equation}
For each $(s,r,o)\in\mathcal{K}$, we denote the reverse of $r$ by $r^{-1}$, and add an additional triplet $(o,r^{-1},s)$ to $\mathcal{K}$.

For an entity $e$, we denote by $N_\mathcal{K}(e)$ its \emph{neighborhood} in $\mathcal{K}$, \ie all related entities with the involved relations.
Formally, 
\begin{equation}
N_\mathcal{K}(e)=\{(r,e')|(e,r,e')\in \mathcal{K}\}\text{.}
\end{equation}
We denote the projection of $N_\mathcal{K}(e)$ on $\mathcal{E}$ and $\mathcal{R}$ by $N_{\mathcal{E}}(e)$ and $N_{\mathcal{R}}(e)$, respectively.
Here $N_{\mathcal{E}}(e)$ are \emph{neighbors} and $N_{\mathcal{R}}(e)$ are \emph{neighboring relations}.
When the context is clear, we simplify the $i$-th entity $e_i$ by its subscript $i$.
We denote vectors by bold lower letters, and matrices or sets of vectors by bold upper letters.

Given a knowledge graph $\mathcal{K}$, we would like to learn a neighborhood aggregator $A$ that acts as follows:
\begin{itemize}
\item For an entity $e_i$ on $\mathcal{K}$, $A$ depends on $e_i$'s neighborhood $N_\mathcal{K}(i)$ to embed $e_i$ as a low-dimensional vector $\mathbf{e}_i$;
\item For an unknown triplet $(s,r,o)$, the embeddings of $s$ and $o$ output by $A$ suggest the plausibility of the triplet.
\end{itemize}
When a new entity emerges with some triplets involving $\mathcal{E}$ and $\mathcal{R}$, we could apply such an aggregator $A$ on its newly established neighborhood, and use the output embedding to infer new facts about it.

\begin{figure}
	\centering
	\includegraphics[scale=0.35]{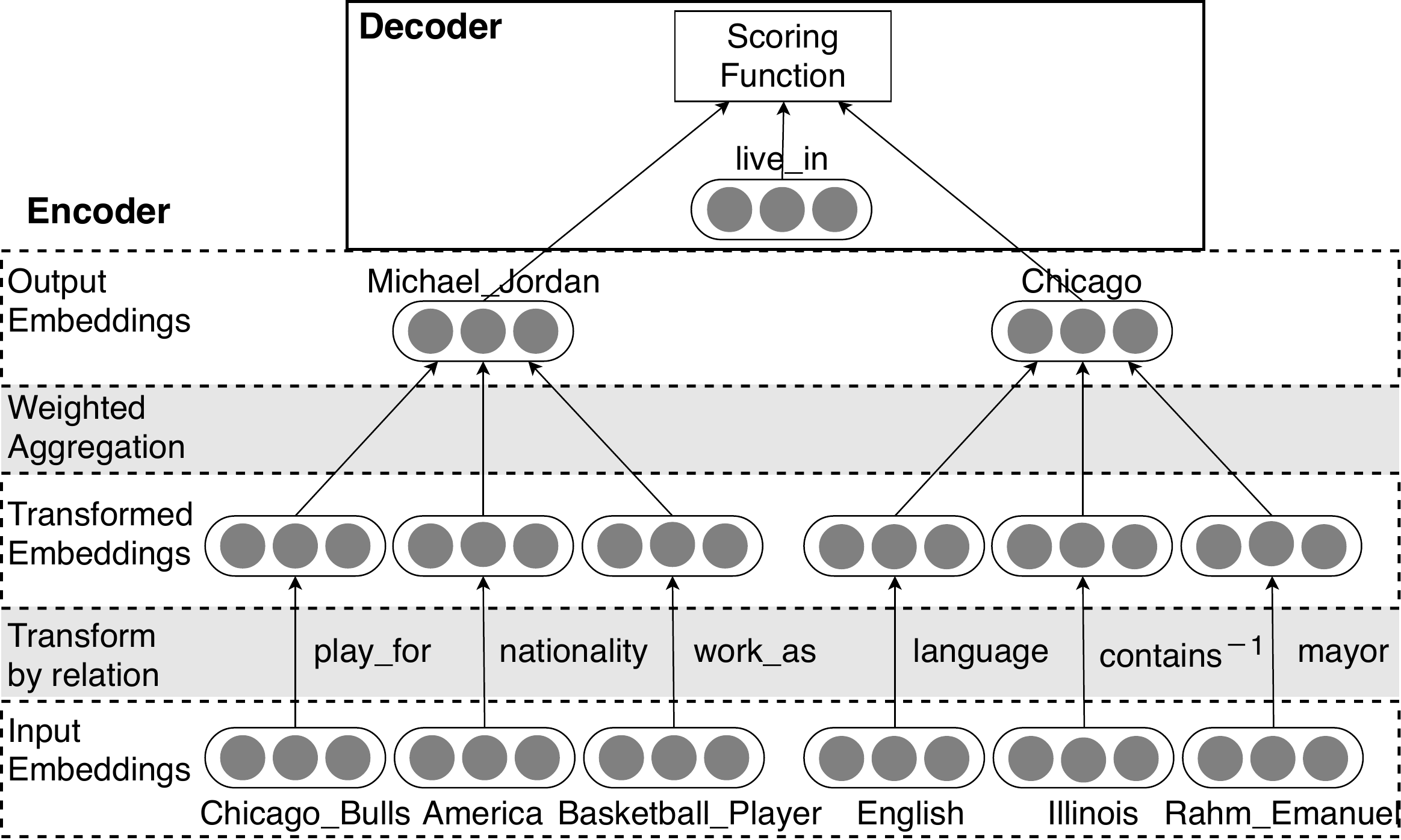}
	\caption{The encoder-decoder framework.} 
	\label{fig:framework}
\end{figure}

\subsection{Framework}

To obtain such a neighborhood aggregator $A$, we adopt an encoder-decoder framework
as illustrated by Figure~\ref{fig:framework}.
Given a training triplet, the encoder $(s,r,o)$ encodes $s$ and $o$ into two embeddings with $A$.
The decoder measures the plausibility of the triplet, and provides feedbacks to the encoder to adjust the parameters of $A$.
In the remainder of this section, we describe general configurations of the two components.

\subsubsection{Encoder}\label{sec:enc}

As specified in Figure~\ref{fig:framework}, for an entity $e_i$ on focus, the encoder works on a collection of input neighbor embeddings, and output $e_i$'s embedding.
To differentiate between input and output embeddings, we use superscripts $I$ and $O$ on the respective vectors.
Let $\mathbf{e}_j^I\in\mathbb{R}^d$, which is obtained from an embedding matrix $\mathbf{W}_e\in\mathbb{R}^{n\times d}$, be the embedding of a neighbor $e_j$, where $(r,e_j)\in N_\mathcal{K}(i)$.
To reflect the impact of relation $r$ on $e_j$, we apply a relation-specific \emph{transforming function} $T_r(.)$ on $\mathbf{e}_j^I$ as follows,
\begin{equation}
	T_r(\mathbf{e}_j^I)=\mathbf{e}_j^I-\mathbf{w}_r^\top\mathbf{e}_j^I\mathbf{w}_r,
\end{equation} 
where $\mathbf{w}_r$ is the transforming vector for relation $r$ and is restricted as a unit vector.
We adopt this transformation from~\citenoun{wang2014knowledge} since it does not involve matrix product operations and is of low computation complexity.

After neighbor embeddings are transformed, these transformed embeddings are fed to the aggregator $A$ to output an embedding $\mathbf{e}_i^O$ for the target entity $e_i$, \ie
\begin{equation}
	\mathbf{e}_i^O=A(\{T_r(\mathbf{e}_j^I)|(r,e_j)\in N_\mathcal{K}(i)\}).
\end{equation}

By definition, an aggregator $A$ essentially takes as input a collection of vectors $\mathbf{X}=\{\mathbf{x}_j\}$ ($\mathbf{x}_j \in\mathbb{R}^d$) and maps them to a single vector.
With this observation, the following two types of functions seem to be natural choices for neighborhood aggregators, and have been adopted previously:
\begin{itemize}
\item\textbf{Pooling Functions.}
A typical pooling function is mean-pooling, which is defined by $A(\mathbf{X})=\frac{1}{|X|}\sum_{j}\mathbf{x}_j$. Besides mean-pooling, other previously adopted choices include sum- and max-pooling~\cite{ijcai2017-250}.
Due to their simple forms, pooling functions are permutation-invariant, but consider the neighbors equally.
It is aware of neither potential redundancy in the neighborhood nor the query relations.

\item\textbf{Recurrent Neural Networks (RNNs).}
In various natural language processing tasks, RNNs prove effective in modeling sequential dependencies.
In~\cite{hamilton2017inductive}, the authors adopt an RNN variant LSTM~\cite{hochreiter1997long} as neighborhood aggregator, \ie $A(\mathbf{X})=\text{LSTM}(\mathbf{x}_1,\dots, \mathbf{x}_{|\mathbf{X}|})$.
To train and apply the LSTM-based aggregator, they have to randomly permute the neighbors, which violates the permutation variance property.
\end{itemize}

\subsubsection{Decoder}\label{sec:dec}

Given the subject and object embeddings $\mathbf{s}^O$ and $\mathbf{o}^O$ output by the encoder, the decoder is required to measure the plausibility of the training triplet.
To avoid potential mixture with relations $r$ in the neighborhood, we refer to the relation in the training triplet by \emph{query relation}, and denote it by $q$ instead.
After looking up $q$'s representation $\mathbf{q}$ from an embedding matrix $\mathbf{W}_r\in\mathbb{R}^{m\times d}$, the decoder scores the training triplet $(s,q,o)$ with a scoring function $\phi^O(s,q,o): \mathbb{R}^d\times \mathbb{R}^d\times \mathbb{R}^d\to\mathbb{R}$.
Following~\citenoun{ijcai2017-250}, we mainly investigate a scoring function based on TransE~\cite{bordes2013translating} defined by
\begin{equation}\label{eq:score_out}
\phi^O(s,q,o)=-\vert \mathbf{s}^O + \mathbf{q} - \mathbf{o}^O\vert_{L1},
\end{equation}
where $|.|_{L1}$ denotes the L1 norm.
To test whether the studied aggregators generalize among different scoring function, we will also consider several alternatives in experiments.

\section{Logic Attention Network}

As discussed above, traditional neighborhood aggregators do not preserve all desired properties.
In this section, we describe a novel aggregator, namely Logic Attention Network (LAN), which addresses all three properties.
We also provide details in training the LAN aggregator.

\subsection{Incorporating Neighborhood Attention}

Traditional neighborhood aggregators only depend on collections of transformed embeddings.
They neglect other useful information in the neighborhood $N_\mathcal{K}(i)$ and the query relation $q$, which may facilitate more effective aggregation of the transformed embeddings.
To this end, we propose generalizing the aggregators from $A(\mathbf{X})$ to $A(\mathbf{X};N_\mathcal{K}(i),q)$.

Specifically, for an entity $e_i$, its neighbors $e_j$ should contribute differently to $\mathbf{e}_i^O$ according to its importance in representing $e_i$.
To consider the different contribution while preserving the permutation invariance property, we employ a weighted or attention-based aggregating approach on the transformed embeddings.
The additional information in $N_\mathcal{K}(i)$ and $q$ is then exploited when estimating the attention weights. Formally, we obtain $\mathbf{e}_i^O$ by
\begin{equation}\label{eq:lan}
\mathbf{e}_i^O=\sum_{(r,e_j)\in N_\mathcal{K}(i)}\alpha_{j|i,q}T_r(\mathbf{e}_j^I).
\end{equation}
Here $\alpha_{j|i,q}$ is the attention weight specified for each neighbor $e_j$ given $e_i$ and the query relation $q$.

To assign larger weights $\alpha_{j|i,q}$ to more important neighbors, from the perspective of $e_i$, we ask ourselves two questions at progressive levels: 1) What types of neighboring relations may lead us to potentially important neighbors? 2) Following those relations, which specific neighbor (in transformed embedding) may contain important information?
Inspired by the two questions, we adopt the following two mechanisms to estimate $\alpha_{j|i,q}$.
\subsubsection{Logic Rule Mechanism} 

Relations in a KG are simply not independent of each other.
For an entity $e$, one neighboring relation $r_1$ may imply the existence of another neighboring relation $r_2$, though they may not necessarily connect $e$ to the same neighbor.
For example, a neighboring relation \textsf{play\_for} may suggest the home city, \ie \textsf{live\_in}, of the current athlete entity.
Following notations in logics, we denote potential dependency between $r_1$ and $r_2$ by a ``logic rule'' $r_1 \Rightarrow r_2$.
To measure the extent of such dependency, we define the \emph{confidence} of a logic rule $r_1 \Rightarrow r_2$ as follows:
	\begin{equation}
		\mathcal{P}(r_1\Rightarrow r_2)=\frac{\sum_{e\in\mathcal{E}}\mathbbm{1}(r_1\in N_\mathcal{R}(e)\land r_2\in N_\mathcal{R}(e))}{\sum_{e\in\mathcal{E}}\mathbbm{1}(r_1\in N_\mathcal{R}(e))}.
	\end{equation}
Here the function $\mathbbm{1}(x)$ equals $1$ when $x$ is true and $0$ otherwise. 
As an empirical statistic over the entire KG, $\mathcal{P}(r_1\Rightarrow r_2)$ is larger if more entities with neighboring relation $r_1$ also have $r_2$ as a neighboring relation.

With the confidence scores $\mathcal{P}(r_1\Rightarrow r_2)$ between all relation pairs at hand, we are ready to characterize neighboring relations $r$ that lead to important neighbors.
On one hand, such a relation $r$ should have a large $\mathcal{P}(r\Rightarrow q)$, \ie it is statistically relevant to $q$.
Following the above example, \textsf{play\_for} should be consulted to if the query relation is \textsf{live\_in}.
On the other hand, $r$ should not be implied by other relations in the neighborhood.
For example, no matter whether the query relation is \textsf{live\_in} or not, the neighboring relation \textsf{work\_as} should not be assigned too much weight, because sufficient information is already provided by \textsf{play\_for}.

Following the above intuitions, we implement the logic rule mechanism of measuring neighboring relations' usefulness as follow:
	\begin{equation}
	\alpha_{j|i,q}^\text{Logic}=\frac{\mathcal{P}(r\Rightarrow q)}{\max(\{\mathcal{P}(r^{'}\Rightarrow r)|r^{'}\in N_\mathcal{R}(e_i)\land r^{'}\neq r\})}.
	\end{equation}
We note that $\alpha_{j|i,q}^\text{Logic}$ promotes relations $r$ strongly implying $q$ (the numerator) and demotes those implied by some other relation in the same neighborhood (the denominator).
In this manner, our logic rule mechanism addresses both query relation awareness and neighborhood redundancy awareness.
	
\subsubsection{Neural Network Mechanism}

With global statistics about relations, the logic rule mechanism guides the attention weight to be distributed at a coarse granularity of relations.
However, it may be insufficient not to consult finer-grained information hidden in the transformed neighbor embeddings to determine which neighbor is important indeed.
To take the transformed embeddings into consideration, we adopt an attention network~\cite{bahdanau2014neural}.

Specifically, given a query relation $q\in\mathcal{R}$, the importance of an entity $e_i$'s neighbor $e_j$ is measured by
	\begin{equation}
    \begin{aligned}
    	\alpha_{j|i,q}^\text{NN} &=\text{softmax}(\alpha^{'}_{j|i,q})=\frac{\exp(\alpha^{'}_{j|i,q})}{\sum_{j^{'}\in N_\mathcal{E}(i)}\exp(\alpha^{'}_{j^{'}|i,q})}. \\
    \end{aligned}
    \end{equation}
Here the unnormalized attention weight $\alpha^{'}_{j|i,q}$ is given by an attention neural network as
	\begin{equation}\label{eq:att}
		\alpha^{'}_{j|i,q}=\mathbf{u}_a^\top\cdot \text{tanh}(\mathbf{W}_a\cdot[\mathbf{z}_q;T_r(\mathbf{e}_j^I)]).
	\end{equation}
In this equation, $\mathbf{u}_a$ and $\mathbf{W}_a\in\mathbb{R}^{d\times 2d}$ are global attention parameters, while $\mathbf{z}_q$ is a relation-specific attention parameter for the query relation $q$.
All those attention parameters are regarded as parameters of the encoder, and learned directly from the data.

Note that, unlike the logic rule mechanism at relation level, the computation of $\alpha_{j|i,q}^\text{NN}$ concentrates more on the neighbor $e_j$ itself.
This is useful when the neighbor entity $e_j$ is also helpful to explain the current training triplet.
For example, in Figure~\ref{fig:framework}, the neighbor \textsf{Chicago\_Bulls} could help to imply the object of \textsf{live\_in} since there are other athletes playing for \textsf{Chicago\_Bulls} while living in \textsf{Chicago}.
Although working at the neighbor level, the dependency on transformed neighbor embeddings $T_r(\mathbf{e}_j^I)$ and the relation-specific parameter $\mathbf{z}_q$ make $\alpha_{j|i,q}^\text{NN}$ aware of both neighborhood redundancy and the query relation.
    
Finally, to incorporate these two weighting mechanisms together in measuring the importance of neighbors, we employ a double-view attention and reformulate Eq.~(\ref{eq:lan}) as
\begin{equation}
\mathbf{e}_i^O=\sum_{(r,e_j)\in N_\mathcal{K}(i)}(\alpha_{j|i,q}^\text{Logic}+\alpha_{j|i,q}^\text{NN})T_r(\mathbf{e}_j^I).
\end{equation}

\subsection{Training Objective}

To train the entire model in Figure~\ref{fig:framework}, we need both positive triplets and negative ones.
All triplets $\mathcal{K}$ from the knowledge graph naturally serve as positive triplets, which we denote by $\Delta$.
To make up for the absence of negative triplets, for each $(s,q,o)\in\Delta$, we randomly corrupt the object or subject (but not both) by another entity in $\mathcal{E}$, and denote the corresponding negative triplets by $\Delta^{'}_{(s,q,o)}$. Formally,
\begin{equation}
\Delta^{'}_{(s,q,o)}=\{(s^{'},q,o)|s^{'}\in\mathcal{E}\}\cup\{(s,q,o^{'})|o^{'}\in\mathcal{E}\}.
\end{equation}  
To encourage the decoder to give high scores for positive triplets and low scores for negative ones, we apply a margin-based ranking loss on each triplet $(s,q,o)$, \ie
\begin{equation}\label{eq:rank_loss}
l^O(s,q,o) =[\gamma -\phi^O(s,q,o)+\phi^O(s^{'},q,o^{'})]_+.
\end{equation}
Here $[x]_+=max\{0,x\}$ denotes the positive part of x, and $\gamma$ is a hyper-parameter for the margin. 
Finally, the training objective is defined by
\begin{equation}\label{eq:object_function}
\min\sum\limits_{(s,q,o)\in\Delta}
\sum\limits_{(s^{'}, q,o^{'})\in\Delta^{'}_{(s,q,o)}}l^O(s,q,o).
\end{equation}
\renewcommand{\arraystretch}{1.0}
\begin{table}[t]\small

    \caption{Statistics of the processed FB15K dataset.}\label{tab:dataset_statistics}
	\begin{tabular}{l|ccc|ccc}
		\toprule
        \multirow{2}{*}{Dataset}&\multirow{2}{*}{$\vert\mathcal{R}\vert$}&\multirow{2}{*}{$\vert\mathcal{E}\vert$}&\multirow{2}{*}{$\vert\mathcal{U}\vert$}& \multicolumn{3}{c}{$\vert N_\mathcal{E}(i)\vert$} \\
		&  & & & min & max & avg  \\
		\midrule
		Subject-5&1,250&12,187&1,460&1 & 7,850&41.5\\
        Object-5&1,182&12,269&1,330&1&6,969&39.4\\
        \midrule
        Subject-10&1,170&10,336&2,082&1 & 5,639&31.6\\
        Object-10&1,126&10,603&1,934&1&5,718&30.9\\
        \midrule
        Subject-15&1,073&8,877&2,342&1 & 5,284&25.5\\
        Object-15&1,057&9,246&2,207&1&4,889&25.4\\
        \midrule
        Subject-20&994&7,765&2,544&1 & 4,485&21.1\\
        Object-20&984&8,219&2,351&1&4,105&21.3\\
        \midrule
        Subject-25&990&6,884&2,666&1 & 3,200&17.7\\
        Object-25&912&7,177&2,415&1&3,580&17.9\\
		\bottomrule
	\end{tabular}
\end{table}


\subsubsection{Subtask on Input Embeddings}

The above training objective only optimizes the output of the aggregator, \ie the output entity embeddings $\mathbf{e}^O$.
The input entity embeddings $\mathbf{e}^I$, however, are not directly aware of the structure of the entire KG.
To make the input embeddings and thus the aggregation more meaningful, we set up a subtask for LAN. 

First, we define a second scoring function, which is similar to Eq.~(\ref{eq:score_out}) except that input embeddings $\mathbf{e}^I$ from $\mathbf{W}_e$ are used to represent the subject and object, \ie
\begin{equation}\label{eq:score_in}
\phi^I(s,q,o)=-\vert \mathbf{s}^I + \mathbf{q} - \mathbf{o}^I\vert_{L1}.
\end{equation}
The embedding of query relation $q\in\mathcal{R}$ is obtained from the same embedding matrix $\mathbf{W}_r$ as in the first scoring function. 
Then a similar margin-based ranking loss $l^I(s,q,o)$ as Eq.~(\ref{eq:rank_loss}) is defined for the subtask. Finally, we combine the subtask with the main task, and reformulate the overall training objective of LAN as
\begin{equation}\label{eq:object_function2}
\min\sum\limits_{(s,q,o)\in\Delta}
\sum\limits_{(s^{'}, q,o^{'})\in\Delta^{'}_{(s,q,o)}}[l^O(s,q,o)+l^I(s,q,o)].
\end{equation}

\section{Experimental Configurations}
We evaluate the effectiveness of our LAN model on two typical knowledge graph completion tasks, \ie link prediction and triplet classification. We compare our LAN with two baseline aggregators, MEAN and LSTM, as described in the Encoder section. MEAN is used on behalf of pooling functions since it leads to the best performance in~\citenoun{ijcai2017-250}. LSTM is used due to its large expressive capability~\cite{hamilton2017inductive}.
\subsection{Data Construction}\label{sec:data}

In both tasks, we need datasets whose test sets contain new entities unseen during training. For the task of triplet classification, we directly use the datasets released by~\citenoun{ijcai2017-250} which are based on WordNet11~\cite{socher2013reasoning}.  Since they do not conduct experiments on the link prediction task, we construct the required datasets based on FB15K~\cite{bordes2013translating} following a similar protocol used in~\citenoun{ijcai2017-250} as follows.

\begin{enumerate}
	\item \emph{Sampling unseen entities}.
Firstly, 
we randomly sample $R=\{5\%, 10\%, 15\%, 20\%, 25\%\}$ of the original testing triplets to form a new test set $\mathcal{T}$ for our inductive scenario (\citenoun{ijcai2017-250} samples $N=\{1000,3000,5000\}$ testing triplets).
Then two different strategies\footnote{Note that we do not employ a third \emph{Both} strategy as in the~\cite{ijcai2017-250}, which adds to $\mathcal{U}^{'}$ the entities appearing as both subject and object in $\mathcal{T}$. This is because when doing link prediction, we only predict the unseen entities' missing relations with the existing entities in $\mathcal{E}$. } are used to construct the candidate unseen entities $\mathcal{U}^{'}$. One is called \emph{Subject}, where only entities appearing as the subjects in $\mathcal{T}$ are added to $\mathcal{U}^{'}$. Another is called \emph{Object}, where only objects in $\mathcal{T}$ are added to $\mathcal{U}^{'}$. For an entity $e\in\mathcal{U}^{'}$, if it does not have any neighbor in the original training set, such an entity is filtered out, yielding the final unseen entity set $\mathcal{U}$. For a triplet $(s,r,o)\in\mathcal{T}$, if $s\in\mathcal{U}\land o\in\mathcal{U}$ or $s\in\mathcal{E}\land o\in\mathcal{E}$, it is removed from $\mathcal{T}$.
	\item \emph{Filtering and splitting data sets}.
	The second step is to ensure that unseen entities would not appear in final training set or validation set. We split the original training set into two data sets, the new training set and auxiliary set. For a triplet $(s,r,o)$ in original training set, if $s,o\in\mathcal{E}$, it is added to the new training set. If $s\in\mathcal{U}\land o\in\mathcal{E}$ or $s\in\mathcal{E}\land o\in\mathcal{U}$, it is added to the auxiliary set, which serves as existing neighbors for unseen entities in $\mathcal{T}$.
Finally, for a triplet $(s,r,o)$ in the original validation set, if $s\in\mathcal{U}$ or $o\in\mathcal{U}$, it is removed from the validation set.
\end{enumerate}
The statistics for the resulting $2\times5=10$ datasets using \emph{Subject} and \emph{Object} strategies are in Table~\ref{tab:dataset_statistics}.

\begin{table}[t]\small
\setlength\tabcolsep{2.5pt}
	\caption{Evaluation accuracy on triplet classification (\%).}\label{tab:tc_result}
	\centering
	\begin{tabular}{c|ccc|ccc|ccc}
		\toprule
        &\multicolumn{3}{c|}{Subject}&\multicolumn{3}{c|}{Object}&\multicolumn{3}{c}{Both} \\
		Model & 1000 & 3000 & 5000 & 1000& 3000& 5000 & 1000& 3000& 5000 \\
		\midrule
		MEAN &87.3&84.3&83.3&84.0&75.2&69.2&83.0&73.3&68.2 \\
		LSTM &87.0&83.5&81.8&82.9&71.4&63.1&78.5&71.6&65.8 \\
		\midrule
		LAN&\textbf{88.8}&\textbf{85.2}&\textbf{84.2}&\textbf{84.7}&\textbf{78.8}&\textbf{74.3}&\textbf{83.3}&\textbf{76.9}&\textbf{70.6}\\
		\bottomrule
	\end{tabular}
\end{table}

\begin{table*}[t]
	\caption{Evaluation results for link prediction.}\label{tab:lp_result}
	\centering
	\small
	\begin{tabular}{c|ccccc|ccccc}
		\toprule
		& \multicolumn{5}{c|}{\textbf{Subject-10}} &  \multicolumn{5}{c}{\textbf{Object-10}} \\ 
		
		Model & MR & MRR & Hits@10 & Hits@3 & Hits@1 & MR & MRR & Hits@10 & Hits@3 & Hits@1 \\
		\midrule
		MEAN&293&0.310&48.0&34.8&22.2& \textbf{353}&0.251&41.0&28.0&17.1 \\
		LSTM&353&0.254&42.9&29.6&16.2& 504&0.219&37.3&24.6&14.3\\
		\midrule
		LAN&\textbf{263}&\textbf{0.394}&\textbf{56.6}&\textbf{44.6}&\textbf{30.2}& 461& \textbf{0.314}&\textbf{48.2}&\textbf{35.7}&\textbf{22.7} \\
		\bottomrule
	\end{tabular}
\end{table*}

\section{Experiments on Triplet Classification}
Triplet classification aims at classifying a fact triplet $(s,r,o)$ as true or false. 
In the dataset of~\citenoun{ijcai2017-250}, triplets in the validation and testing sets are labeled as true or false, while triplets in the training set are all true ones.

To tackle this task, we preset a threshold $\delta_r$ for each relation r. If $\phi^O(s,r,o) \ge \delta_r$, the triplet is classified as positive, otherwise it is negative. We determine the optimal $\delta_r$ by maximizing classification accuracy on the validation set.

\subsection{Experimental Setup}
Since this task is also conducted in~\citenoun{ijcai2017-250}, we use the same configurations with learning rate $\alpha=0.001$, embedding dimension $d=100$, and margin $\gamma=300.0$ for all datasets. We randomly sample 64 neighbors for each entity. 
Zero padding is used when the number of neighbors is less than 64. 
L2-regularization is applied on the parameters of LAN. The regularization rate is $0.001$.

\subsection{Evaluation Results}
The results are reported in Table~\ref{tab:tc_result}. Since we did not achieve the same results for MEAN as reported in~\citenoun{ijcai2017-250} with either our implementation or their released source code, the best results from their original paper are reported.
From the table, we observe that, on one hand, LSTM results in poorer performance compared with MEAN, which involves fewer parameters though. This demonstrates the necessity of the permutation invariance for designing neighborhood aggregators for KGs. On the other hand, our LAN model consistently achieves the best results on all datasets, demonstrating the effectiveness of LAN on this KBC task.

\begin{table}
\caption{Effectiveness of logic rules on Subject-10.}\label{tab:lr_result}
	\centering
	\small
	\begin{tabular}{c|cccc}
		\toprule
		Model & MRR & Hits@10 &Hit@3& Hits@1 \\
		\midrule
        MEAN&0.310&48.0&34.8&22.2 \\
       	\midrule
        Global-Attention &0.331 & 49.7 &37.7& 24.0 \\
		Query-Attention & 0.355 & 51.9 &39.5& 27.0 \\
        Logic Rules Only&0.375&54.7&42.7&28.0 \\
        LAN &\textbf{0.394}&\textbf{56.6}&\textbf{44.6}& \textbf{30.2} \\
		\bottomrule
	\end{tabular}
\end{table}

\section{Experiments on Link Prediction}
Link prediction in the inductive setting aims at reasoning the missing part ``?'' in a triplet when given $(s, r, ?)$ or $(?, r, o)$ with emerging entities $s$ or $o$ respectively. To tackle the task, we firstly hide the object (subject) of each testing triplet in Subject-R (Object-R) to produce a missing part. 
Then we replace the missing part with all entities in the entity set $\mathcal{E}$ to construct candidate triplets.
We compute the scoring function $\phi^O(s,r,o)$ defined in Eq.~(\ref{eq:score_out}) for all candidate triplets, and rank them in descending order. 
Finally, we evaluate whether the ground-truth entities are ranked ahead of other entities.
We use traditional evaluation metrics as in the KG completion literature, \ie Mean Rank (MR), Mean Reciprocal Rank (MRR), and the proportion of ground truth entities ranked top-k (Hits@k, $k\in\{1,3,10\}$). Since certain candidate triplets might also be true, we follow previous works and filter out these fake negatives before ranking. 

\subsection{Experimental Setup}
We search the best hyper-parameters of all models according to the performance on validation set. In detail, we search learning rate $\alpha$ in $\{0.001,0.005,0.01,0.1\}$,  embedding dimension for neighbors $d$ in $\{20, 50,100, 200\}$, and margin $\gamma$ in $\{0.5, 1.0, 2.0, 4.0\}$. The optimal configurations are $\alpha=0.001, d=100, \gamma=1.0$ for all the datasets. 

\begin{table}[t]
\caption{Different scoring functions on Subject-10.}\label{tab:scoring_result}
	\centering
	\small
	\begin{tabular}{c|c|cccc}
		\toprule
		Encoder & Decoder & MRR & Hits@10 & Hit@3 & Hits@1 \\
		\midrule
		MEAN& Distmult & 0.297&45.8&33.5&21.2 \\
        MEAN& Complex &0.286&44.7&32.2&20.4 \\
        MEAN& Analogy &0.242&38.3&26.5&17.1 \\
        MEAN& TransE &0.310&48.0&34.8&22.2 \\
        \midrule
		LAN& Distmult & 0.378&53.4&43.2&29.3 \\
        LAN& Complex & 0.371&53.1&42.2&28.7 \\
        LAN & Analogy & 0.375&53.2&42.6&29.2 \\
        LAN& TransE & \textbf{0.394}&\textbf{56.6}&\textbf{44.6}&\textbf{30.2} \\
		\bottomrule
	\end{tabular}
\end{table}

\subsection{Experimental Results}
\begin{figure*}[t]
\parbox{.235\linewidth}{
	\includegraphics[width=\linewidth]{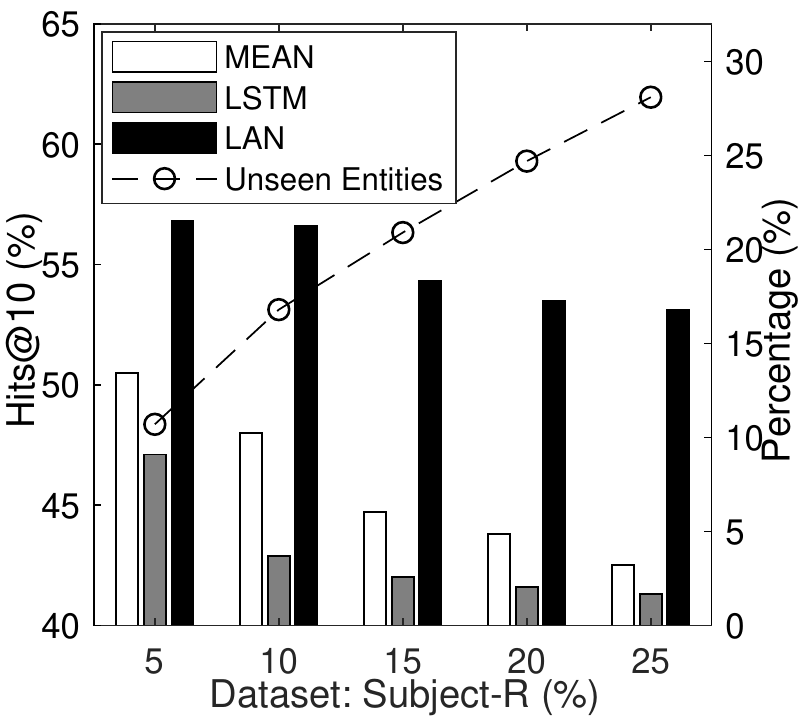}
    \includegraphics[width=\linewidth]{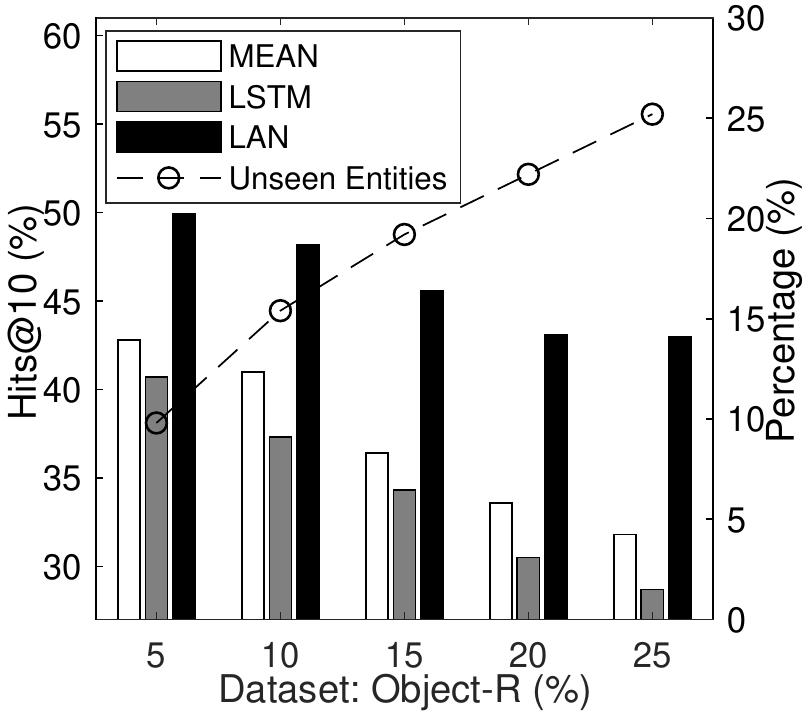}
    \caption{Results on Subject-R and Object-R.}
    \label{fig:unseen}
    }
    \quad
\parbox{.74\linewidth}{
	\centering
	\includegraphics[width=\linewidth]{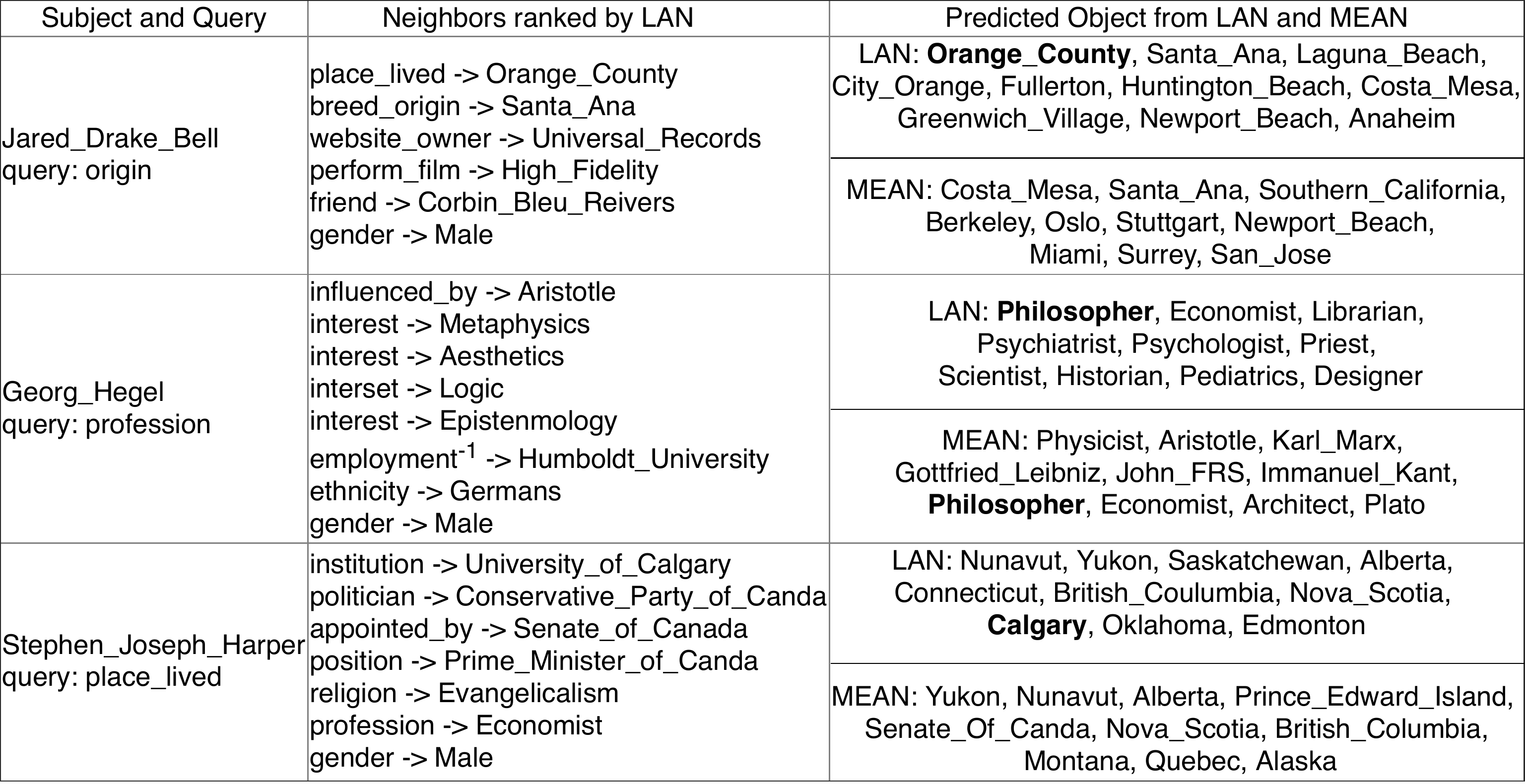}
	\captionof{table}{The sample cases. The left column contains the emerging entity and the query relation. The middle column contains the neighbors ranked in a descending order according to the weights specified by LAN. The right column contains the ranked prediction from LAN and MEAN. The correct predictions are marked in bold.}
	\label{fig:weight}
    }
\end{figure*}

The results on Subject-10 and Object-10 are reported in Table~\ref{tab:lp_result}. The results on other datasets are similar and we summarize them later in Figure~\ref{fig:unseen}.  From Table~\ref{tab:lp_result}, we still observe consistent results for all the models as in the triplet classification task. Firstly, LSTM results in the poorest performance on all datasets. Secondly, our LAN model outperforms all the other baselines significantly, especially on the Hit@k metrics. The improvement on the MR metric of LAN might not be considerable. This is due to the flaw of the MR metric since it is more sensitive to lower positions of the ranking, which is actually of less importance. The MRR metric is proposed for this reason, where we could observe consistent improvements brought by LAN. The effectiveness of LAN on link prediction validates LAN's superiority to other aggregators and the necessities to treat the neighbors differently in a permutation invariant way. To analyze whether LAN outperforms the others for expected reasons and generalizes to other configurations, we conduct the following studies.

\subsubsection{Necessity of Query Relation Awareness}
In this experiment, we would like to confirm that it's necessary for the aggregator to be aware of the query relation. Specifically, we investigate the attention neural network and design two degenerated baselines. One is referred to as Query-Attention and is simply an attention network as in LAN except that the logic rule mechanism is removed. The other is referred to as Global-Attention, which is also an attention network except that the query relation embedding $\mathbf{z}_q$ in Eq.~(\ref{eq:att}) is masked by a zero vector. The results are reported in Table~\ref{tab:lr_result}. We observe that although superior to MEAN, Global-Attention is outperformed by Query-Attention, demonstrating the necessity of query relation awareness.
The superiority of Global-Attention over MEAN could be attributed to the fact that the attention mechanism is effective to identify the neighbors which are globally important regardless of the query.

\subsubsection{Effectiveness of Logic Rule Mechanism}
We find that the logic rules greatly help to improve the attention network in LAN. We confirm this point by conducting further experiments where the logic rule mechanism is isolated as a single model (referred to as Logic Rules Only). The results are also demonstrated in Table~\ref{tab:lr_result},
from which we find that Query-Attention outperforms MEAN by a limited margin. Meanwhile, Logic Rules Only outperforms both MEAN and Query-Attention by significant margins. These results demonstrate the effectiveness of logic rules in assigning meaningful weights to the neighbors. Specifically, in order to generate representations for unseen entities, it is crucial to incorporate the logic rules to train the aggregator, instead of depending solely on neural networks to learn from the data. By combining the logic rules and neural networks, LAN takes a step further in outperforming all the other models.

\subsubsection{Generalization to Other Scoring Functions}
To find out whether the superiority of LAN to the baselines can generalize to other scoring functions, we replace the scoring function in Eq.~(\ref{eq:score_out}) and Eq.~(\ref{eq:score_in}) by three typical scoring functions mentioned in Related Works.
We omit the results of LSTM, for it is still inferior to MEAN. The results are listed in Table~\ref{tab:scoring_result}, from which we observe that with different scoring functions, LAN outperforms MEAN consistently by a large margin on all the evaluation metrics. Note that TransE leads to the best results on MEAN and LAN. 

\subsubsection{Influence of the Proportion of Unseen Entities}
It's reasonable to suppose that when the ratio of the unseen entities over the training entities increases (namely the observed knowledge graph becomes sparser), all models' performance would deteriorate. To figure out whether our LAN could suffer less on sparse knowledge graphs, we conduct link prediction on datasets with different sample rates $R$ as described in Step 1 of the Data Construction section. 
The results are displayed in Figure~\ref{fig:unseen}. We observe that the increasing proportion of unseen entities certainly has a negative impact on all models. However, the performance of LAN does not decrease as drastically as that of MEAN and LSTM, indicating that LAN is more robust on sparse KGs. 

\subsection{Case Studies on Neighbors' Weights}
In order to visualize how LAN specifies weights to neighbors, we sample some cases from the Subject-10 testing set. From Table~\ref{fig:weight}, we have the following observations. First, with the query relation, LAN could attribute higher weights to neighbors with more relevant relations. In the first case, when the query is \textsf{origin}, the top two neighbors are involved by \textsf{place\_lived} and \textsf{breed\_origin}, which are helpful to imply \textsf{origin}. In addition, in all three cases, neighbors with relation \textsf{gender} gain the lowest weights since they imply nothing about the query relation. Second, LAN could attribute higher weights to neighbor entities that are more informative. When the query relation is \textsf{profession}, the neighbors \textsf{Aristotle}, \textsf{Metaphysics} and \textsf{Aesthetics} are all relevant to the answer \textsf{Philosopher}. In the third case, we also observe similar situations. Here, the neighbor with the highest weight is \textsf{(institution, University\_of\_Calgary)} since the query relation \textsf{place\_lived} helps the aggregator to focus on the neighboring relation \textsf{institution}, then the neighbor entity \textsf{University\_of\_Calgary} assists in locating the answer \textsf{Calgary}. 

\section{Conclusion}
In this paper, we address inductive KG embedding, which helps embed emerging entities efficiently.
We formulate three characteristics required for effective neighborhood aggregators. To meet the three characteristics, we propose LAN, which attributes different weights to an entity's neighbors in a permutation invariant manner, considering both the redundancy of neighbors and the query relation. The weights are estimated from data with logic rules at a coarse relation level, and neural attention network at a fine neighbor level. Experiments show that LAN outperforms baseline models significantly on two typical KG completion tasks.

\section{Acknowledgements}
We thank the three anonymous authors for their constructive comments.
This work is supported by the National Natural Science Foundation of China (61472453, U1401256, U1501252, U1611264, U1711261, U1711262).

\bibliographystyle{aaai}
\bibliography{aaai19}

\end{document}